\documentclass[final]{cvpr}

\usepackage{times}
\usepackage{epsfig}
\usepackage{graphicx}
\usepackage{verbatim}
\graphicspath{{images/}}
\usepackage{subcaption}

\usepackage{amssymb}
\usepackage{latexsym}
\usepackage{amsmath}
\usepackage{float} 

\DeclareMathOperator*{\argmax}{arg\,max}

\usepackage[pagebackref=true,breaklinks=true,colorlinks,bookmarks=false]{hyperref}

\usepackage{diagbox}

\makeatletter
\robustify\@latex@warning@no@line
\makeatother
\usepackage{authblk}

\def\confYear{CVPR 2021}
\begin{document}

\title{Out-of-Distribution Detection for Dermoscopic Image Classification}

\author[1, 2]{Mohammadreza Mohseni\thanks{Corresponding Author}}
\author[2]{Jordan Yap}
\author[2]{William Yolland}
\author[2]{Majid Razmara}
\author[1, 3]{M Stella Atkins}
\affil[1]{School of Computing Science, Simon Fraser University}
\affil[2]{MetaOptima Technology Inc}
\affil[3]{Department of Skin Science and Dermatology, University of British Columbia}
\affil[ ] {\vspace{-3mm}}
\affil[ ]{ \tt\small\{mmohseni, stella\}@sfu.ca, \{jordan, william, majid\}@metaoptima.com}

\maketitle

\begin{abstract}
Medical image diagnosis can be achieved by deep neural networks, provided there is enough varied training data for each disease class. However, a hitherto unknown disease class not encountered during training will inevitably be misclassified, even if predicted with low probability. 
This problem is especially important for medical image diagnosis, when an image of a hitherto unknown disease is presented for diagnosis, especially when the images come from the same image domain, such as dermoscopic skin images.

Current out-of-distribution detection algorithms act unfairly when the in-distribution classes are imbalanced, by favouring the most numerous disease in the training sets. This could lead to false diagnoses for rare cases which are often medically important.

We developed a novel yet simple method to train neural networks, which enables them to classify in-distribution dermoscopic skin disease images and also detect novel diseases from dermoscopic images at test time. 
We show that our BinaryHeads model not only does not hurt classification balanced accuracy when the data is imbalanced, but also consistently improves the balanced accuracy. We also introduce an important method to investigate the effectiveness of out-of-distribution detection methods based on presence of varying amounts of out-of-distribution data, which may arise in real-world settings.

\end{abstract}


\section{Introduction}
It is important to diagnose malignant skin lesions early. In particular, early detection and surgical treatment of malignant melanoma can result in excellent patient outcomes \cite{mel}. 
Other malignant skin lesions, including basal cell carcinoma (BCC) and squamous cell carcinoma (SCC), can also be threatening if left untreated \cite{nonmel}.
However, differentiating malignant skin lesions from benign skin lesions such as nevi and seborrheic keratoses (SK) is often difficult even for trained clinicians.
Several clinical algorithms have been developed to aid clinicians make a diagnosis of a skin lesion, such as the visual clues to help to diagnose malignant melanoma. These clinical algorithms include the well-known ABCD criteria (Asymmetry, Border irregularity, Colour irregularity, Diameter) \cite{ABCDNachbar, ABCDAbbasi}, and the 7-point check-list \cite{SevenPointArgenziano, SevenPointBetta}.
But even with these clues, Heal \etal showed it can be extremely difficult for physicians to make a differential diagnosis between the commonly-encountered skin lesions, such as nevi and non-melanocytic lesions like seborrheic keratoses (SK), and rarely-encountered malignant melanomas \cite{heal}.

A magnified dermoscopic view taken very close to, or in contact with, the skin is frequently used to supplement the observational clinical view for diagnosis of skin lesions \cite{braun2005}. A dermoscopic image, typically taken at magnification 10-15 times and with polarized lighting, can show skin details at much higher resolution, including dermoscopic structures a few millimeters under the surface of the skin including textures such as pigment networks, dots, globules and streaks \cite{braun2005, dermmel}.

\begin{figure}[htp]
\centering
\begin{subfigure}{0.45\linewidth}
\includegraphics[width=0.9\linewidth]{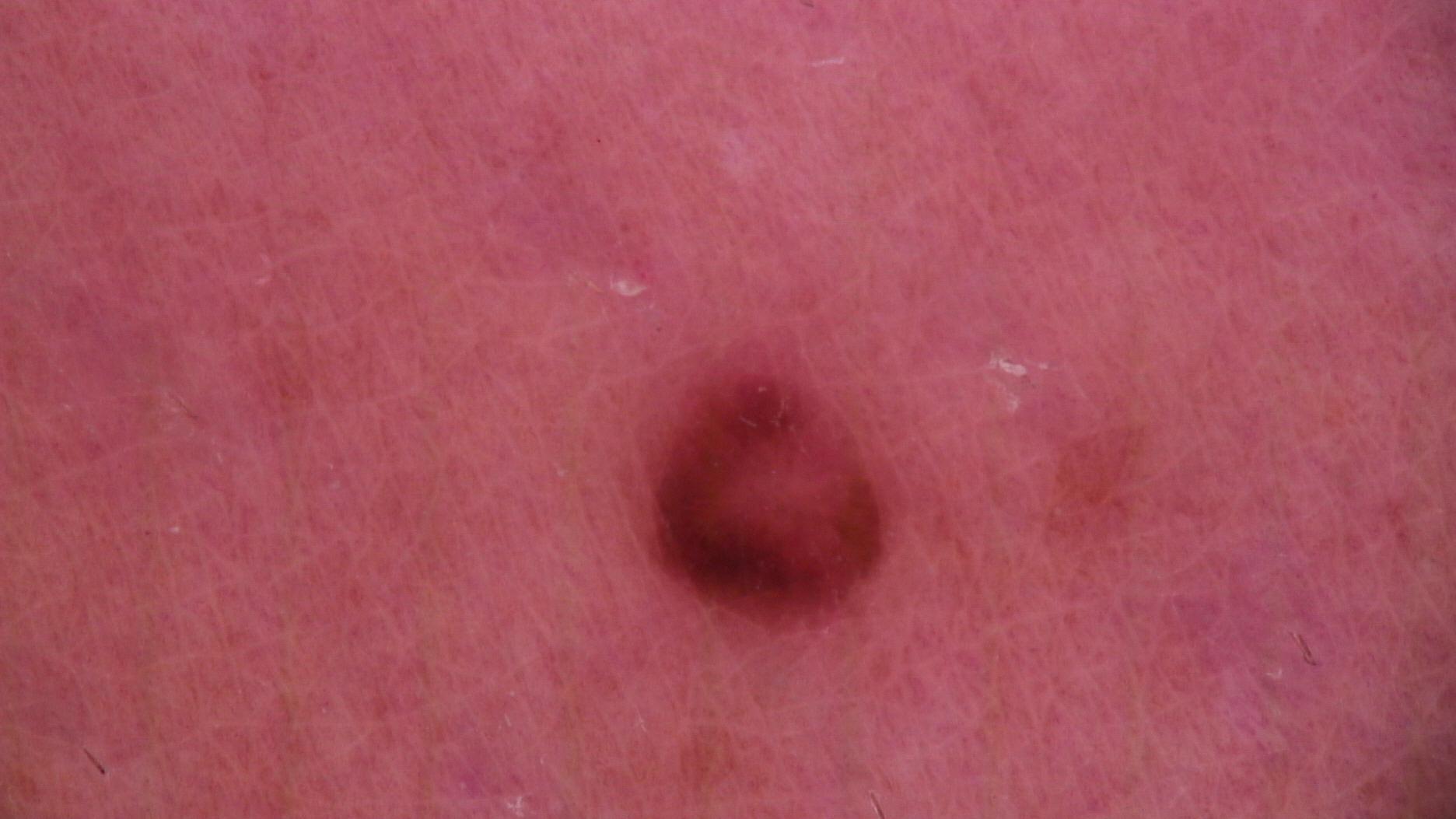} 
\caption{Dermoscopic nevus}
\label{fig:dermnevus}
\end{subfigure}
\begin{subfigure}{0.45\linewidth}
\includegraphics[width=0.9\linewidth]{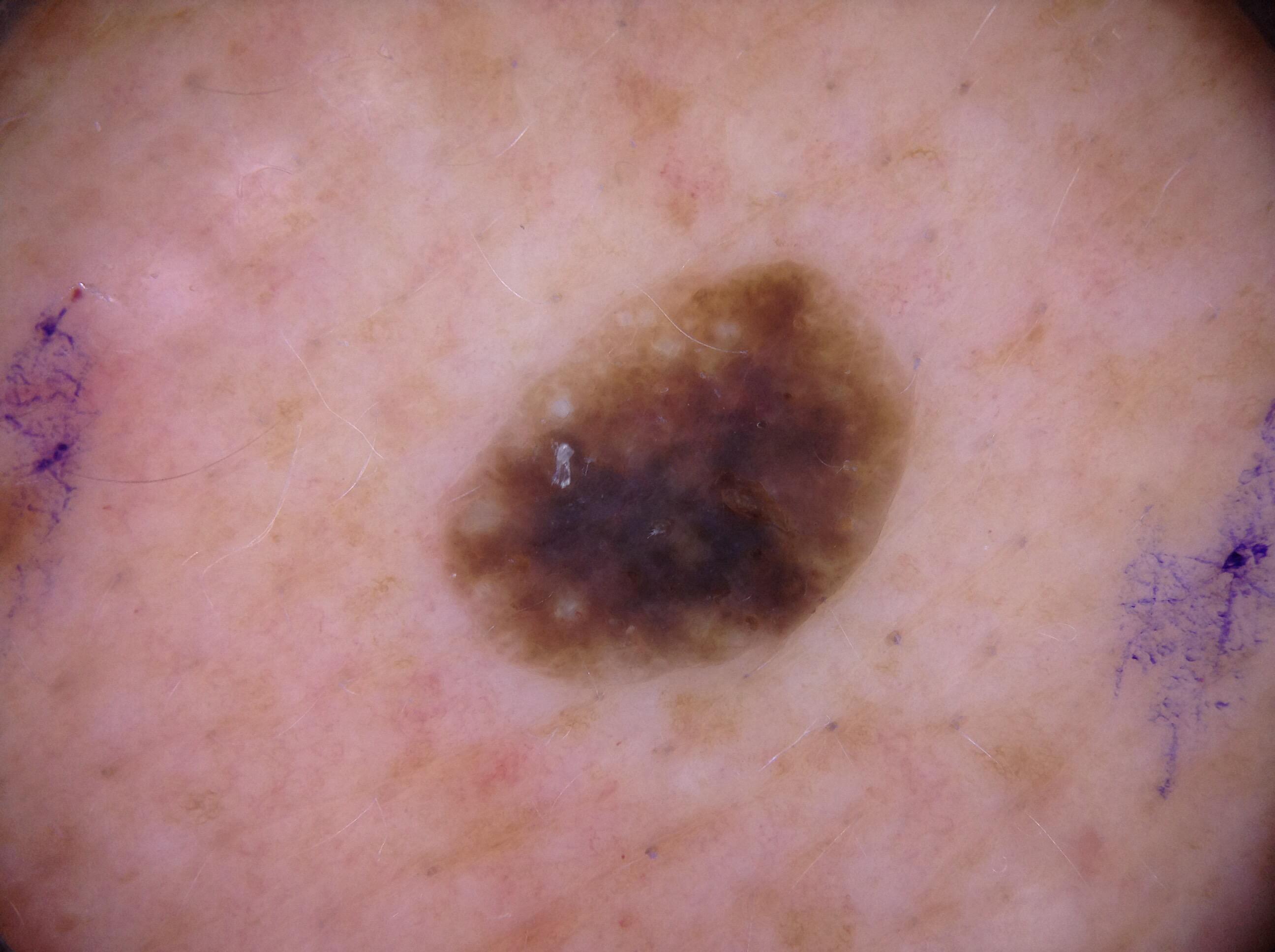}
\caption{Dermoscopic SK}
\label{fig:dermmel}
\end{subfigure}

\caption{Dermoscopic images showing how SK can mimic a nevus (publicly available skin images from ISIC 2020 dataset \cite{rotemberg2021patient}).}
\label{fig:derm}
\end{figure}

In contrast with the previously mentioned clinical algorithms, a separate set of clues for diagnosis from dermoscopic images has also been developed such as the dermoscopy ABCD rule where "D" stands for dermoscopic structures instead of "diameter" as it does for the clinical ABCD observation \cite{dermrule}.
These dermoscopy rules are primarily intended to help identify malignant melanoma, but the differential diagnosis remains very challenging, with many possible features to identify and use \cite{msk}. Inter-class similarities and intra-class dissimilarities are examples of what make differential diagnoses challenging. For example, benign seborrheic keratoses (SK) mimic SCCs, BCCs, and malignant melanomas, especially for patients with many atypical (dysplastic) nevi \cite{SoenksenUD2021}. 
There are also many intra-class dissimilarities, where a given disease may have many subtypes according to colour and texture \cite{dermmel}.
Dermoscopic images of two nevi and SKs are shown in Fig \ref{fig:derm}, illustrating the difficulty in differentiating between these two diagnoses.

Using dermoscopy images to supplement the clinical view improves diagnosis \cite{argenziano}, \cite{msk}. However, it is very important to provide adequate training in dermoscopy for primary care physicians to make good diagnoses \cite{koelink, dermtraining, jones}, and even then it is very difficult to make correct differential diagnoses. There is also a lot of uncertainty in diagnosis, which means that many more benign lesions are excised than are strictly necessary \cite{welch2021}.

We were motivated to develop an automated tool to help dermatologists and primary care physicians to perform differential diagnosis of dermoscopic skin lesions encountered in the clinic. Since developing an algorithm which is able to detect every skin disease's type is not possible, we built a tool which is able to express uncertainty when presented with a previously unseen disease. Our algorithm can be an aid in treatment decisions and detection of lesions which are not in the commonly encountered categories.


We also show that current out-of-distribution (OOD) detection algorithms face many challenges in our evaluation setting. These algorithms act well when the OOD images are coming from a different domain, but they act poorly when the OOD images share many geometric and semantic features with the in-distribution images. Also, we show that classes with smaller number of images in the training set contribute more to false positives in the OOD detection.\\

Our algorithms are designed to perform multi-class classification of dermoscopic images of skin lesions with high accuracy, and with an important additional out-of-distribution (OOD) class by leveraging a separate binary classifier for each in-distribution class.\\

\begin{figure*}[t!]
\centering
\includegraphics[width=\linewidth]{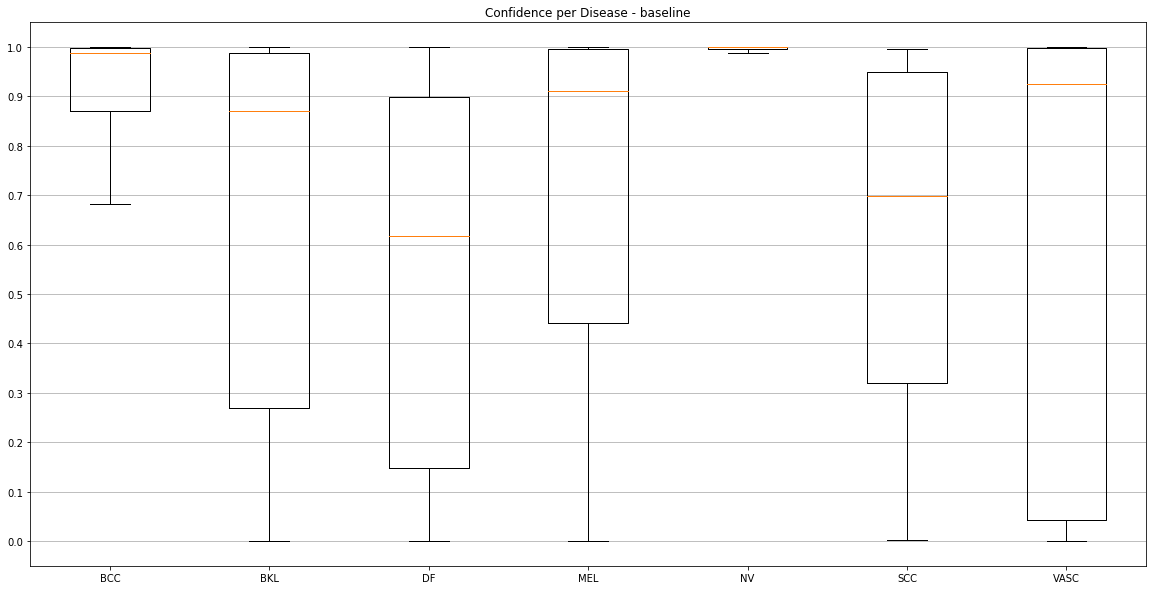} 
\caption{Range of softmax probabilities predicted for each class. Some classes tend to have higher range of confidence compared to others.}
\label{fig:softmax_motivation}
\end{figure*}

\section {Related Work}
\subsection {Automated analysis of dermoscopic skin images}
\label{related}
Research on automated analysis of dermoscopic images of skin lesions initially used image processing methods, often focusing on the "ABCD" features such as border irregularity \cite{border2003}, or texture features \cite{pn, streaks}.

Recently deep learning approaches have proven very successful, despite having been initially hamstrung by a lack of data \cite{paul2011}.
However with much more labelled data, Esteva \etal showed deep learning models could perform as well as expert dermatologists \cite{Esteva2017}. 
Furthermore, Yap \etal showed that incorporating metadata provided even better results \cite{yap}.

The interest in deep learning methods has been catalyzed by challenges hosted by the International Skin Imaging Collaboration (ISIC), which has released thousands of high-quality images to the public.

In 2018, Tschandl \etal published the HAM10000 dataset \cite{ham10000} with over 10,000 labelled dermoscopic images, which were used for the 2018 ISIC skin diagnosis challenge.
The main task was to perform differential diagnosis of 7 classes of skin lesions using the HAM10000 dataset.
The possible 7 disease categories were: melanoma (MEL), melanocytic nevus (NV), basal cell carcinoma (BCC), actinic keratosis/Bowens disease (AKIEC), benign keratosis (BKL) (solar lentigo/ seborrheic keratosis / lichen planus-like keratosis), dermatofibroma (DF), and vascular (VASC).

The winning results showed balanced multi-class accuracy of 88.5\%, a commendable result given the highly imbalanced nature of the data in each class. An important subsequent study by Tschandl \etal showed that the top three algorithms actually out-performed dermatologists with respect to most outcome measures \cite{lancet}. 
However, as Tschandl \etal noted, a limitation of these
algorithms is their decreased performance for out-of-distribution images, which should be addressed in future research \cite{lancet}.

The main task in the 2019 ISIC challenge \cite{isic2019} was to classify dermoscopic images among 9 different diagnosis categories: the 7 classes from the 2018 challenge where the AK class now specified only actinic keratosis (AK) as opposed to (AKIEC), plus Squamous cell carcinoma (SCC), and a class representing "none of the others" ie out of distribution.
For the ISIC 2019 challenge, 25,331 images were available for training across the 8 known different categories. 

The winning balanced multiclass accuracy was 63.6\% with the addition of external training data, and 60.7\% without using external data.

This much lower accuracy reflected the addition of two extra classes in the test set (SCC and the "other" class).

For both of the 2018 and 2019 challenges, the classes had highly imbalanced distributions, reflecting the distribution of lesions biopsied in a real clinical setting.

The 2020 ISIC challenge task was to identify melanoma in images of skin lesions \cite{isic2020}.
The leaderboard shows there were over 3300 entries, with the winning score of 94.9\% (area under the ROC curve).
Of note is the fact that in 2020 there were 33k training images, but only 1.8\% were malignant (vs 17.8\% in 2019, 10 times more) ie the data was much more strongly imbalanced compared to previous years.

A summary of recent AI approaches, mostly published in 2017-2019, to diagnosing skin lesions is given by Goyal \etal \cite{goyal2020}.

\subsection{Out-of-Distribution Detection Algorithms}
Despite all the advances in classification, AI models still have difficulty in expressing their uncertainty and detecting OOD samples. Gal \etal estimated uncertainty in deep neural networks using Monte-Carlo DropOut \cite{DropoutUncertainty}. They proposed that measuring uncertainty can be an effective way to detect OOD samples ("special cases"). They provided a mathematically grounded proof on why averaging multiple stochastic forward passes can capture uncertainty in dropout neural networks. Der Kiureghian and Ditlevsen suggested breaking uncertainty into two types: Aleatoric and Epistemic \cite{UncertaintyTypes}. Aleatoric uncertainty comes from the uncertainty in the data. An image of a lesion which is hard to tell if it is a benign melanocytic nevus or malignant melanoma is an example of one which might accompany high aleatoric uncertainty. Epistemic uncertainty comes from a lack of predictive power, due to observing only a subset of the true data distribution at training time. 
In our work, by restricting the domain of novel disease images to dermoscopic images, aleatoric uncertainty is increased considerably. We show that even in the scenario when there is higher aleatoric uncertainty between in-distribution samples and OOD samples, detecting OOD lesion images is still possible.\\

Hendrycks and Gimpel proposed an initial baseline method for OOD detection which works based on thresholding softmax probabilities \cite{baseline}. Bendale and Boult proposed a methodology for unknown class detection at test time by introducing an OpenMax model layer \cite{OSDN}. Liang \etal showed that temperature scaling and adding small perturbations to the input image are two effective tools in separating in-distribution samples and OOD samples \cite{ODIN}. Lee \etal calculated a confidence score based on Mahalanobis distance between a given sample and class-conditional probability distributions \cite{mahalanobis}. Vyas \etal identified OOD samples by enforcing a margin between in-distribution and OOD samples softmax probabilities in the loss function across an ensemble of classifiers \cite{metaEnsemble}. Later, Hendrycks \etal showed that incorporating auxiliary OOD images during the training stage and enforcing the model to produce a uniform response on auxiliary OOD samples can boost a model's uncertainty estimation at test time \cite{OutlierExposure}. Recently, Liu \etal showed that energy-based models associate higher energy values with OOD samples \cite{energy}. They identified OOD samples by thresholding a samples' energy calculated over the output of the model. They also provided a fine-tuning method to improve their results even more in the setting where auxiliary OOD samples are available at training time. Shafaei \etal discussed different points of views when tackling the OOD detection problem \cite{Shafaei-OD-Test}. They also proposed a three-dataset evaluation scheme for more reliable evaluation of OOD detection algorithms.\\

Another limitation of current outlier detection algorithms is that they aim to sacrifice the model's balanced accuracy in order to maximize accuracy in outlier detection. These models look at the OOD detection problem as a binary classification problem. However, with the existence of underrepresented classes in skin diseases, these methods act harshly on underrepresented classes. Fig \ref{fig:softmax_motivation} shows the variation of predicted probabilities per samples in different classes. It can be seen that defining a global threshold on the probabilities only takes the false positive samples from underrepresented classes. This consequently decreases a model's balanced accuracy significantly.

\subsection{Out-of-Distribution Detection in Dermatology}
There has also been some research in order to detect OOD samples in the context of skin disease detection. As mentioned above, ISIC 2019 introduced a challenge \cite{isic2019} where training data contained 8 classes.
However, in the testing phase there was an "Other" class which did not belong to the training classes. 

Pacheco \etal used the Gram matrix to estimate a given sample's deviation and identify various types of OOD samples from different domains \cite{Pacheco}. Their results showed that OOD detection on domains closer to in-distribution data is much more challenging than on other domains. Combalia \etal \cite{combalia} used test data augmentation and Monte-Carlo test time dropout (\cite{DropoutUncertainty}) to measure both aleatoric and epistemic uncertainty at the same time. Bagchi \etal utilized a two-level ensembling technique to classify skin lesions and detect novel classes \cite{SkinMetaEnsemble}. They also showed that detecting outliers results in a decrease in balanced multi-class accuracy measured on the ISIC 2019 test set.

\section{Data}
In order to train and evaluate our algorithms, we used publicly available dermoscopic images from the ISIC 2019 \cite{ham10000, codella, bcn20000}
and ISIC 2020 challenge \cite{rotemberg2021patient}.

From the ISIC 2020 challenge, we only used the data with labels present in the ISIC 2019 classes or where the label could be mapped to one of the classes in ISIC 2019 challenge. So images labelled as lichenoid keratosis, solar lentigo, and seborrheic keratosis were labelled as benign keratosis.

We treated actinic keratosis as the out-of-distribution class and trained our model on the rest of the classes. We split the data such that $80\%$ of lesion IDs of in-distribution data went to the training set, for a total of 26,400 training images. The rest of the data (in-distribution and out-of-distribution) was split equally between the validation and test set. More details can be found in Table \ref{table:class_counts}.\\

\begin{table}[H]
\centering
\begin{tabular}{| c | c | c | c |} 
 \hline
 Class Name & Training & Validation & Test \\ [0.5ex] 
 \hline
 BCC & 2807 (11\%) & 299 (12\%) & 217 (9\%)\\ [0.5ex]
 \hline
 BKL & 2397 (9\%) & 259 (10\%) & 147 (6\%)\\ [0.5ex]
 \hline
 DF & 187 (1\%) & 28 (1\%) & 24 (1\%)\\ [0.5ex]
 \hline
 MEL & 4720 (18\%) & 193 (8\%) & 193 (8\%)\\ [0.5ex]
 \hline
 NV & 15609 (59\%) & 1230 (48\%) & 1229 (53\%)\\ [0.5ex]
 \hline
 SCC & 496 (2\%) & 79 (3\%) & 53 (2\%)\\ [0.5ex]
 \hline
 VASC & 184 (1\%) & 40 (2\%) & 29 (1\%)\\ [0.5ex]
 \hline
 AK & 0 (0\%) & 428 (17\%) & 439 (19\%)\\ [0.5ex]
 \hline
\end{tabular}
\caption{Number of images per class used and the percentage per set.}
\label{table:class_counts}
\end{table}

\begin{figure*}[!t]
\centering
\includegraphics[width=0.85\linewidth]{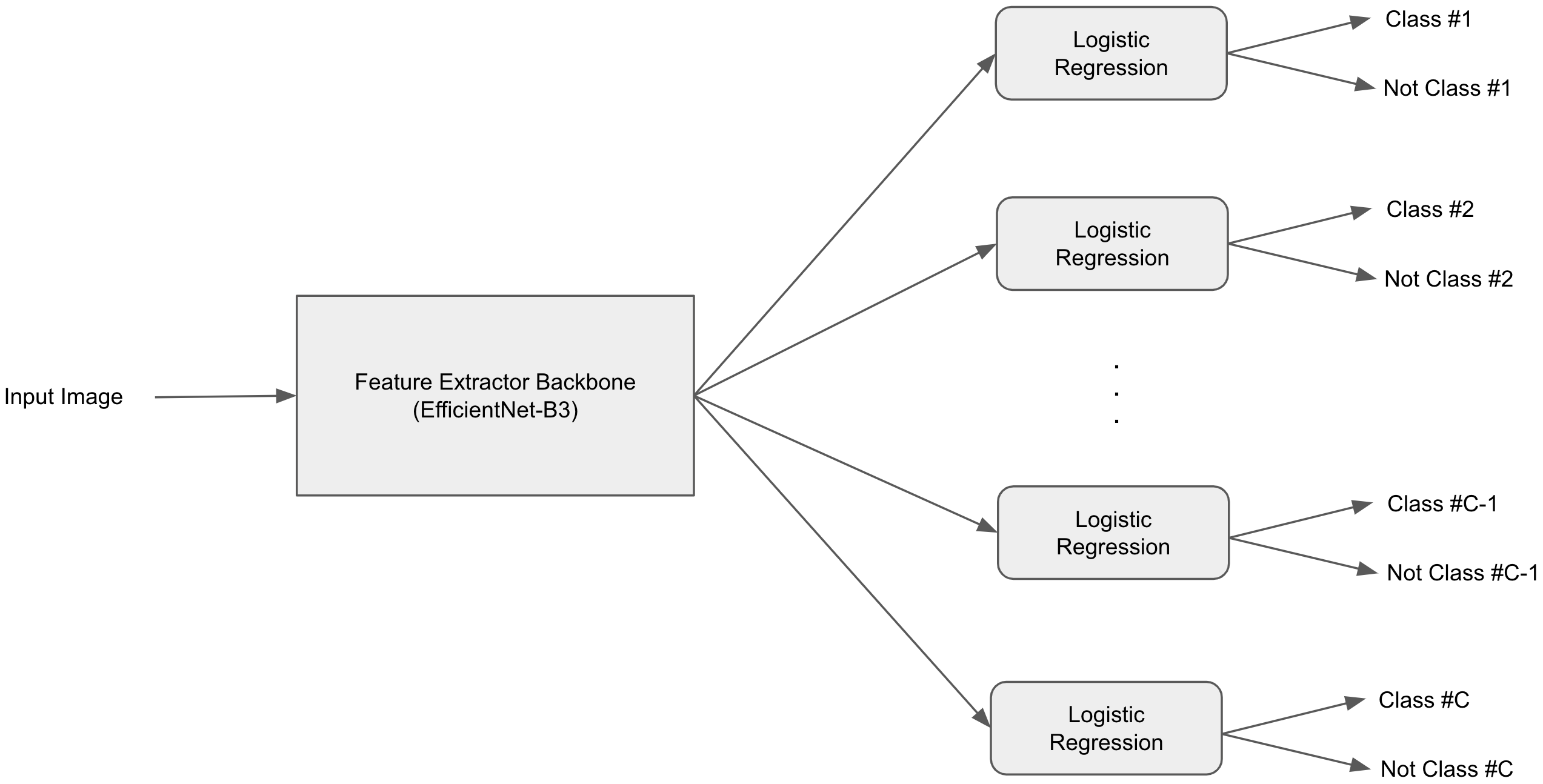}
\caption{BinaryHeads network}
\label{fig:BH-Network}
\end{figure*}

\section{Method}
One less-travelled path in multi-class classification is using $C$ one-versus-the-rest classifiers where $C$ is the number of classes. Although this approach comes with difficulties in training, it can be shown to effectively identify OOD samples. Each of the $C$ classifiers is trained to predict whether the given sample belongs to its corresponding class. If no classifier accepts the given sample, the sample will be classified as OOD.

In order to perform this task, we propose a simple yet flexible neural network architecture called BinaryHeads (BH) which transforms training multiple classifiers, into a multi-task learning framework. A BH network consists of a shared feature extractor backbone and multiple classification heads, each corresponding to one known class in the training set. In our experiments we used EfficientNet-B3 as feature extractor backbone and used a simple logistic regression unit for each classification head. An overview of our BH model can be seen in Fig \ref{fig:BH-Network}.

\subsection{Training}
When training the BH network, each dermoscopic image belongs to exactly one class. We set the ground truth for the corresponding classification head to be 1 and the rest should predict 0. Using this schema we train all classification heads together. Each binary head backpropagates through its logistic regression unit to the backbone network at the same time.

\subsection{Inference}
At inference time, we define $C$ thresholds, each corresponding to one of the classification classes in the training set. In order to get the network's prediction on each sample, we first discard the classes where the sample probability is lower than the class threshold. Among remaining classes, we then choose the class with the highest probability, or we predicted the sample to be OOD if no class remains.
\begin{equation}
\begin{split}
confidence = \max_{i=1}^{|C|} H(prob_i-threshold_i)*prob_i \\
candidateClass = \argmax_{i=1}^{|C|} H(prob_i-threshold_i)*prob_i \\
prediction= 
\begin{cases}
    candidateClass & \text{if } confidence > 0\\ & \\
    OOD              & \text{otherwise}
\end{cases}
\end{split}
\label{eq:prediction}
\end{equation}
Equation \ref{eq:prediction} shows the criteria for choosing the predicted class. Note that H is heaviside step function.
The intuition behind having per-class thresholds is that different classes have different amounts of training data and distinctiveness. This leads to some classes being overconfident and others being underconfident as shown in Fig \ref{fig:softmax_motivation}. Hence these thresholds are defined to be more fair to underrepresented classes. It is important to note setting per-class threshold here is possible since predicted probabilities at each head are independent. This means that independent heads can have independent thresholds.

In order to find the most suitable per-class thresholds, we define an initial set of thresholds (can be any arbitrary initial value between 0 and 1). Then by local optimization of these thresholds we maximize balanced accuracy on the validation set. In the local optimization process, each time we randomly choose a disease, and then set its threshold to the value which maximizes the balanced accuracy. We repeat this process until all the diseases have their threshold set and we converge to a local minimum. We find that this method strikes a good balance between complexity and effectiveness.\\

\begin{figure*}[t!]
\centering
\includegraphics[width=0.85\linewidth]{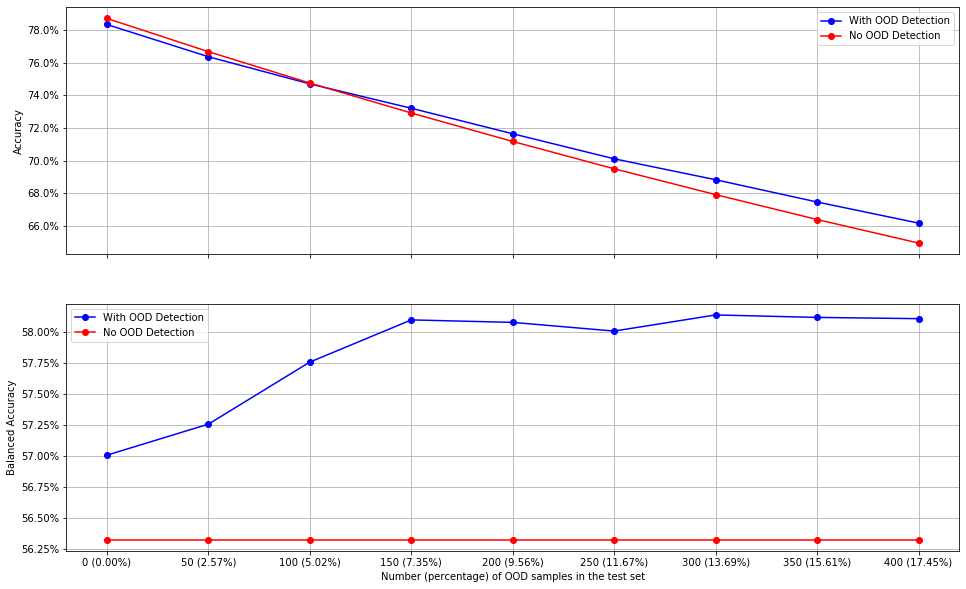}
\caption{Comparing a BH model with and without OOD detection capability. Note that balanced accuracy is calculated across all in-distribution and OOD classes, even when no OOD is in the test set, in which case we assume 0\% sensitivity for OOD class)}
\label{fig:Experiment1}
\end{figure*}

\section{Experiments}
\subsection{Setting}
Some additional training details which helped our models converge were regimes of weighted sampling and data augmentation. These were mainly employed to minimize the effect of class imbalance as much as possible. We augmented the data using random rotation, color jitter, and center crop.\\
We trained the models using stochastic gradient descent as the optimizer and decreased the learning rate on plateau during the training process.
More information on our training hyperparameters can be found in the projects' Github repository \footnote{\url{https://github.com/mrm-196/Derm-OOD}}.

\subsection{Results}
In order to effectively evaluate an OOD detection algorithm in an imbalanced setting, one needs to show how the presence of differing amounts of OOD data in the dataset affects accuracy and balanced accuracy of the system. This is critical, since the number of novel diseases seen in different clinics can vary greatly and it is difficult to impossible to have an accurate estimation of number of OOD samples across different real world scenarios. Thus, it is more realistic to evaluate the performance of the system against various amounts of OOD data present in the test set. Also, the class imbalance in the in-distribution classes suggests the use of a metric which is robust to class imbalance. To evaluate the performance of our classifier at each scenario, we measure both the accuracy of the classifier as well as its balanced accuracy.\\

In practice, a strong consideration for the incorporation of an OOD detection component within a classification framework is its overall effect on classification performance. One needs to show that the inclusion of an OOD detection algorithm does not hurt general performance of the system so much as to render it noticeably less useful. In order to do such evaluation, we compare a BH model which has OOD detection capability, against a vanilla BH model. The vanilla BH model classifies samples by choosing the class with the maximum probability as the prediction and does not label any samples as OOD. Fig \ref{fig:Experiment1} shows the outcome of this experiment across different number of OOD samples in the dataset.\\

\begin{figure*}[!t]
\centering
\includegraphics[width=0.85\linewidth]{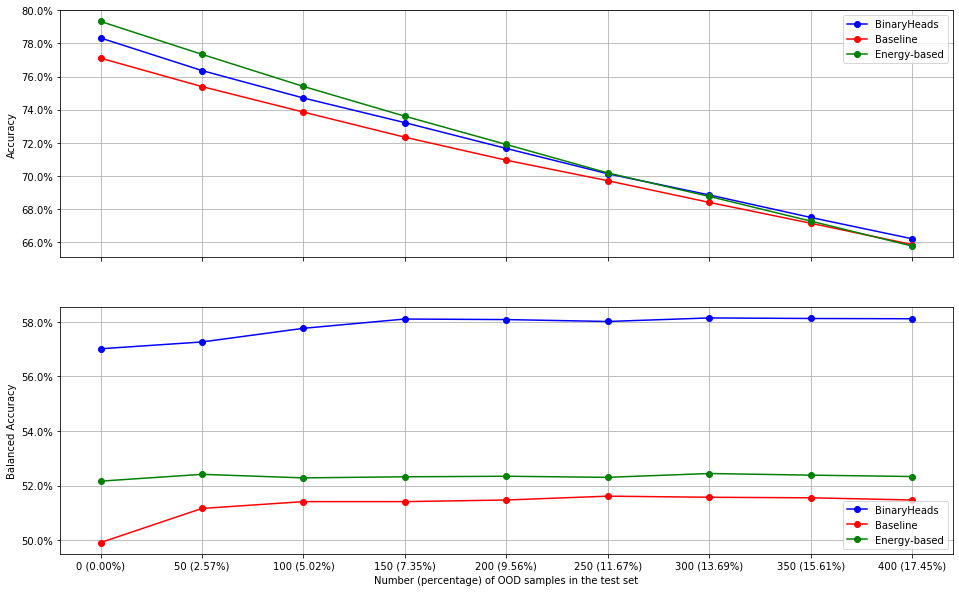}
\caption{Comparing three different systems with three different OOD detection abilities}
\label{fig:Experiment2}
\end{figure*}

\begin{figure}[h]
\centering
\includegraphics[width=\linewidth]{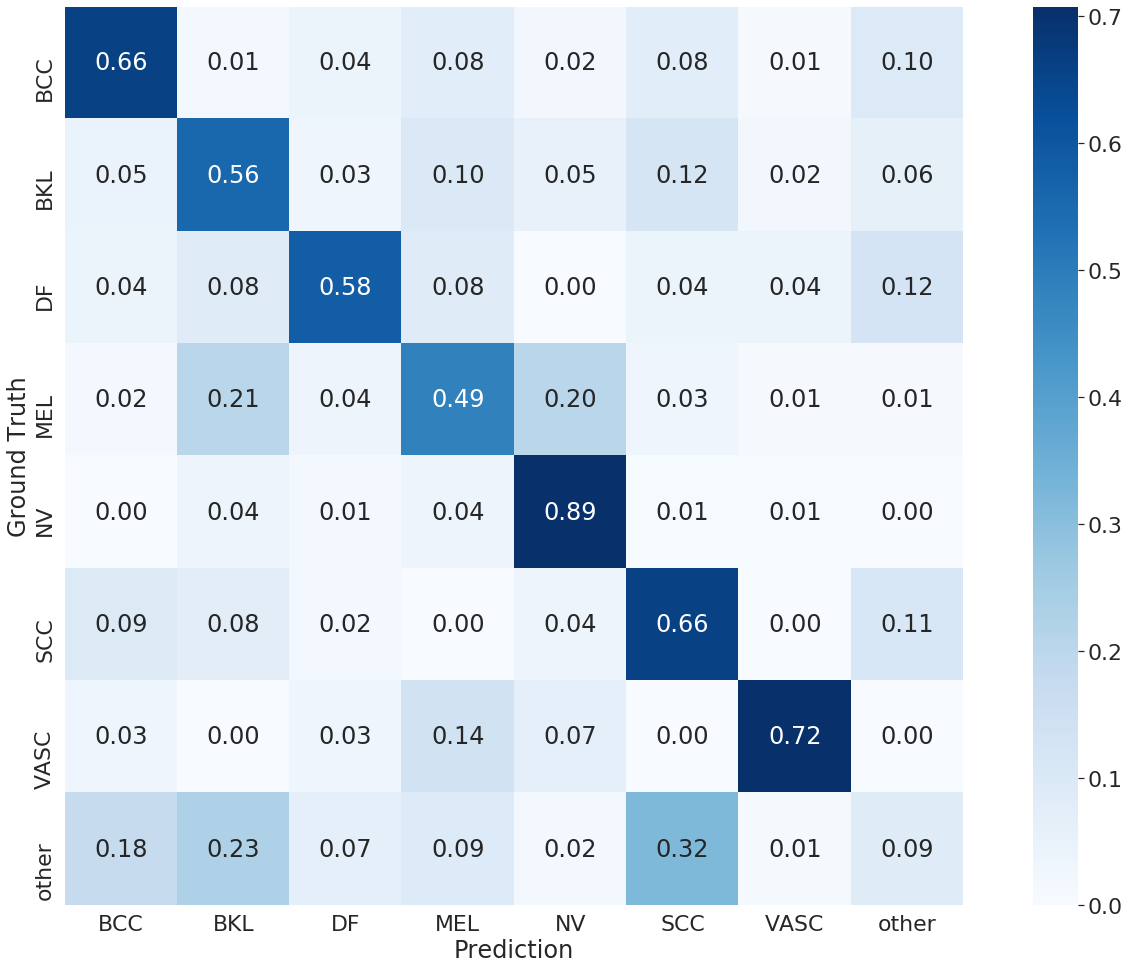}
\caption{Confusion Matrix of BH model on the full test set.}
\label{fig:confusion_matrix}
\end{figure}

We also compared our work against other recent OOD detection algorithms. Of the algorithms listed in related works, we chose the baseline OOD detection algorithm \cite{baseline} and the energy based OOD detection \cite{energy}. The energy-based method currently holds the state of the art in the OOD detection task. Fig \ref{fig:Experiment2} shows a comparison of our method, against the baseline OOD detector, and the energy-based OOD detector. The baseline method identifies a sample as OOD if the maximum class probability is lower than a predefined threshold across all classes. The energy based method initially calibrates the model using temperature scaling before calculating the energy for each sample. Samples with energy higher than a given threshold are then predicted as OOD.
 
Our BH model's accuracy on the full test set was 65.21\% and had a balanced accuracy of 58.10\%. The confusion matrix in Fig \ref{fig:confusion_matrix} shows there is no significant unfairness towards underrepresented classes.

\section{Discussion}
According to our experiments shown in Fig \ref{fig:Experiment1}, the per-class thresholding technique not only helps with identifying new diseases, but also decreases the confusion between in-distribution classes. This can be seen in the scenario when there is no OOD available and we gain 0.68\% balanced accuracy by having per-disease thresholds.\\

The result shows that when the number of OOD samples in the test set is relatively high, our per-class thresholding technique increases both the model's accuracy as well as its balanced accuracy. However, when there are not many OOD samples in the test set, our method hurts accuracy while increasing balanced accuracy.\\

Also, an interesting observation is that current OOD detection approaches seem poorly equipped to detect OOD data which comes from the same domain as the in-distribution classes. This result is also supported by Pacheco \etal \cite{Pacheco}. The reason that we limit our focus to this type of OOD is its importance in the clinical setting. Consequently, as shown in Fig \ref{fig:Experiment2}, our method outperforms the baseline and the energy-based methods in terms of balanced accuracy while having very competitive accuracy in the presence of various amounts of OOD samples.\\

Another observation which can be made from Figures \ref{fig:Experiment1} and \ref{fig:Experiment2} is the strong performance of the BH model even without the ability to detect OOD samples compared to the baseline model. While conventional knowledge suggests that training binary classifiers to detect each class in a one-vs-rest setting is not the best way to train classifiers, we show that such a classifier outperforms its more popular softmax-based cousin. Additionally, we believe that BH model is more flexible and interpretable than a softmax classifier since the BH model can propose multiple candidate classes for the prediction even if those classes are underrepresented or complex to learn.\\

Many OOD detection approaches rely on a set of OOD samples in advance in order to find appropriate thresholds. Another strong point about using BH network in this setting is that setting thresholds can be done even when no OOD samples are available. We experimented with setting thresholds both with and without using OOD samples. Where no OOD samples were used, performance decreased by only 0.26\% accuracy and 0.37\% balanced accuracy which is relatively small and provides a good trade-off to collecting a set of OOD examples, which may not always be feasible.\\

We should mention that although this approach comes with almost no additional costs compared to softmax methods. Training BH networks can be challenging and in some cases they may not converge to a good solution. These networks are similar to multi-task networks and they might benefit from techniques like GradNorm \cite{gradnorm} in order to be trained properly. Future research is suggested on this matter to reveal the true potential of BH networks.

\section{Conclusion}
Well-instructed primary care physicians (PCPs) are able to consistently capture dermoscopic images from a similar domain as training images, rendering classification of dermoscopic lesions feasible. However, training on every possible class of skin diseases is not possible, because some diseases are extremely rare. 
This suggests a need to identify a usable OOD detection method which is trained on in-domain data (common diseases) and is able to identify novel diseases, without largely sacrificing model performance, nor having to be trained/tuned on a large held-out set of OOD samples.

We have presented a novel method for detecting OOD classes and have demonstrated how it might be used to help partition medical images even from within the same domain, such as dermoscopy skin lesion images, as being known or unknown diseases (unseen during training). Importantly, our method does not rely on training with labelled images of every possible skin disease, nor does it rely on having access to out of distribution samples to tune thresholds.

Future work will examine the performance of our method when multiple different disease classes are removed from the training set, and when new disease classes such as some inflammatory conditions are added.
Ultimately, we plan to evaluate the performance of our system through comparison with dermatologists.

\section*{Acknowledgments}
The reserach is funded by the natural sciences and engineering reserach council  (NSERC) OF Canada, grant no. 611106.



\bibliographystyle{ieee_fullname}
\bibliography{egbib}

\begin{thebibliography}{10}\itemsep=-1pt

\bibitem{ABCDAbbasi}
Naheed~R Abbasi, Helen~M Shaw, Darrell~S Rigel, Robert~J Friedman, William~H
  McCarthy, Iman Osman, Alfred~W Kopf, and David Polsky.
\newblock Early diagnosis of cutaneous melanoma: revisiting the abcd criteria.
\newblock {\em JAMA}, 292(22):2771–2776, 2004.

\bibitem{SevenPointArgenziano}
Giuseppe Argenziano, Gabriella Fabbrocini, Paolo Carli, Vincenzo~De Giorgi,
  Elena Sammarco, and Mario Delfino.
\newblock Epiluminescence microscopy for the diagnosis of doubtful melanocytic
  skin lesions: comparison of the abcd rule of dermatoscopy and a new 7-point
  checklist based on pattern analysis.
\newblock {\em Archives of Dermatology}, 134(12):1563–1570, 1998.

\bibitem{argenziano}
G Argenziano, S Puig, and I Zalaudek.
\newblock {\em J Clin Oncol}, 24:1877–1882, 2006.

\bibitem{SkinMetaEnsemble}
Subhranil Bagchi, Anurag Banerjee, and Deepti~R Bathula.
\newblock Learning a meta-ensemble technique for skin lesion classification and
  novel class detection.
\newblock In {\em Proceedings of the IEEE/CVF Conference on Computer Vision and
  Pattern Recognition Workshops}, pages 746--747, 2020.

\bibitem{OSDN}
Abhijit Bendale and Terrance~E Boult.
\newblock Towards open set deep networks.
\newblock In {\em Proceedings of the IEEE conference on computer vision and
  pattern recognition}, pages 1563--1572, 2016.

\bibitem{SevenPointBetta}
G. Betta, Giuseppe~DI Leo, Gabriella Fabbrocini, Alfredo Paolillo, and M.
  Scalvenzi.
\newblock Automated application of the 7-point checklist diagnosis method for
  skin lesions: Estimation of chromatic and shape parameters.
\newblock {\em Proceedings of the IEEE Instrumentation and Measurement
  Technology Conference}, page 1818–1822, 2005.

\bibitem{braun2005}
RP Braun, HS Rabinovitz, M Oliviero, AW Kopf, and JH Saurat.
\newblock Dermoscopy of pigmented skin lesions.
\newblock {\em J Am Acad Dermatol.}, 52(1):109--121, 2005.

\bibitem{nonmel}
Freddie Bray, Jacques Ferlay, Isabelle Soerjomataram, Rebecca~L. Siegel,
  Lindsey~A. Torre, and Ahmedin Jemal.
\newblock Global cancer statistics 2018: Globocan estimates of incidence and
  mortality worldwide for 36 cancers in 185 countries.
\newblock {\em CA: A Cancer journal for Clinicians}, 68:394--424, 2018.

\bibitem{mel}
Cancer.
\newblock Types of cancer: skin cancer: {C}ancer {C}ouncil {A}ustralia.
\newblock 2021.
\newblock Accessed: 2021-03-27.

\bibitem{gradnorm}
Zhao Chen, Vijay Badrinarayanan, Chen-Yu Lee, and Andrew Rabinovich.
\newblock Gradnorm: Gradient normalization for adaptive loss balancing in deep
  multitask networks.
\newblock In {\em International Conference on Machine Learning}, pages
  794--803. PMLR, 2018.

\bibitem{codella}
Noel C.~F. Codella, David Gutman, M.~Emre Celebi, Brian Helba, Michael~A.
  Marchetti, Stephen~W. Dusza, Aadi Kalloo, Konstantinos Liopyris, Nabin
  Mishra, Harald Kittler, and Allan Halpern.
\newblock Skin lesion analysis toward melanoma detection 2018: A challenge
  hosted by the international skin imaging collaboration ({ISIC}).
\newblock 2018.

\bibitem{bcn20000}
Marc Combalia, Noel~CF Codella, Veronica Rotemberg, Brian Helba, Veronica
  Vilaplana, Ofer Reiter, Cristina Carrera, Alicia Barreiro, Allan~C Halpern,
  Susana Puig, et~al.
\newblock Bcn20000: Dermoscopic lesions in the wild.
\newblock {\em arXiv preprint arXiv:1908.02288}, 2019.

\bibitem{combalia}
M. Combalia, F. Hueto, S. Puig, J. Malvehy, and V. Vilaplana.
\newblock Uncertainty estimation in deep neural networks for dermoscopic image
  classification.
\newblock {\em IEEE/CVF Conference on Computer Vision and Pattern Recognition
  Workshops (CVPRW), Seattle, WA, USA}, pages 3211--3220, 2020.

\bibitem{dermtraining}
V De~Bedout, NM Williams, AM Muñoz, AM Londoño, M Munera, N Naranjo, LM
  Rodriguez, AM Toro, F Miao, T Koru-Sengul, and N. Jaimes.
\newblock Skin cancer and dermoscopy training for primary care physicians: A
  pilot study.
\newblock {\em Dermatol Pract Concept.}, 11(1):e2021145, 2021.

\bibitem{UncertaintyTypes}
Armen Der~Kiureghian and Ove Ditlevsen.
\newblock Aleatory or epistemic? does it matter?
\newblock {\em Structural safety}, 31(2):105--112, 2009.

\bibitem{dermrule}
Dermoscopedia.
\newblock 2021.
\newblock Accessed: 2021-03-27.

\bibitem{Esteva2017}
Andre Esteva, Brett Kuprel, Roberto~A. Novoa, Justin Ko, Susan~M. Swetter,
  Helen~M. Blau, and Sebastian Thrun.
\newblock Dermatologist-level classification of skin cancer with deep neural
  networks.
\newblock {\em Nature}, 542(7639):115--118, Feb. 2017.

\bibitem{DropoutUncertainty}
Yarin Gal and Zoubin Ghahramani.
\newblock Dropout as a bayesian approximation: Representing model uncertainty
  in deep learning.
\newblock In Maria~Florina Balcan and Kilian~Q. Weinberger, editors, {\em
  Proceedings of The 33rd International Conference on Machine Learning},
  volume~48 of {\em Proceedings of Machine Learning Research}, pages
  1050--1059, New York, New York, USA, 20--22 Jun 2016. PMLR.

\bibitem{goyal2020}
Manu Goyal, Thomas Knackstedt, Shaofeng Yan, and Saeed Hassanpour.
\newblock Artificial intelligence-based image classification methods for
  diagnosis of skin cancer: Challenges and opportunities.
\newblock {\em Comp. Bio. Med}, 2020.

\bibitem{heal}
C.F. Heal, B.A. Raasch, P.G. Buettner, and D. Weedon.
\newblock Accuracy of clinical diagnosis of skin lesions.
\newblock {\em British Journal of Dermatology}, 2008.

\bibitem{baseline}
D. {Hendrycks} and K. {Gimpel}.
\newblock A baseline for detecting misclassified and out-of-distribution
  examples in neural networks.
\newblock In {\em ICLR}, 2018.

\bibitem{OutlierExposure}
Dan Hendrycks, Mantas Mazeika, and Thomas Dietterich.
\newblock Deep anomaly detection with outlier exposure.
\newblock {\em arXiv preprint arXiv:1812.04606}, 2018.

\bibitem{isic2019}
{ISIC}.
\newblock {International Skin Imaging Consortium Challenge}.
\newblock 2019.
\newblock Accessed: 2021-03-27.

\bibitem{isic2020}
{ISIC}.
\newblock {International Skin Imaging Consortium Challenge}.
\newblock 2020.
\newblock Accessed: 2021-03-27.

\bibitem{dermmel}
Natalia Jaimes, Ashfaq~A. Marghoob, Harold Rabinovitz, Ralph~P. Braun, Alan
  Cameron, Cliff Rosendahl, Greg Canning, and Jeffrey Keir.
\newblock Clinical and dermoscopic characteristics of melanomas on nonfacial
  chronically sun-damaged skin.
\newblock {\em Journal of the American Academy of Dermatology}, 72
  (6):1027--1035, 2015.

\bibitem{jones}
OT Jones, LC Jurascheck, MA van Melle, S Hickman, NP Burrows, PN Hall, J
  Emeery, and FM Walter.
\newblock Dermoscopy for melanoma detection and triage in primary care: a
  systematic review.
\newblock {\em BMJ Open}, 9, 2019.

\bibitem{koelink}
CJ Koelink, KM Vermeulen, BJ Kollen, de~Bock G.H., Dekker J.H., Jonkman M.F.,
  and van der~Heide W.K.
\newblock Diagnostic accuracy and cost-effectiveness of dermoscopy in primary
  care: a cluster randomized clinical trial.
\newblock {\em J Eur Acad Dermatol Venereol.}, 28(11):1442--1449, 2014.

\bibitem{mahalanobis}
Kimin Lee, Kibok Lee, Honglak Lee, and Jinwoo Shin.
\newblock A simple unified framework for detecting out-of-distribution samples
  and adversarial attacks.
\newblock {\em NEURIPS}, 2018.

\bibitem{border2003}
Tim~K. Lee, D McLean, and M.~Stella Atkins.
\newblock Irregularity index: a new border irregularity measure for cutaneous
  melanocytic lesions.
\newblock {\em Medical image analysis}, 7 (1):47--64, 2003.

\bibitem{ODIN}
Shiyu Liang, Yixuan Li, and R. Srikant.
\newblock Enhancing the reliability of out-of-distribution image detection in
  neural networks.
\newblock 2018.

\bibitem{energy}
W. {Liu}, J. {Owens}, X. {Wang}, and Y. {Li}.
\newblock Energy-based out-of-distribution detection.
\newblock In H. Larochelle, M. Ranzato, R. Hadsell, M.~F. Balcan, and H. Lin,
  editors, {\em Advances in Neural Information Processing Systems}, volume~33,
  pages 21464--21475. Curran Associates, Inc., 2020.

\bibitem{ABCDNachbar}
Franz Nachbar, Wilhelm Stolz, Tanja Merkle, Armand~B. Cognetta, Thomas Vogt,
  Michael Landthaler, Peter Bilek, Otto Braun-Falco, and Gerd Plewig.
\newblock The abcd rule of dermatoscopy: high prospective value in the
  diagnosis of doubtful melanocytic skin lesions.
\newblock {\em J of American Academy of Derm.}, 30(4):551--559, 1994.

\bibitem{Pacheco}
A.~G.~C. {Pacheco}, C.~S. {Sastry}, T. {Trappenberg}, S. {Oore}, and R.~A.
  {Krohling}.
\newblock On out-of-distribution detection algorithms with deep neural skin
  cancer classifiers.
\newblock pages 3152--3161, 2020.

\bibitem{rotemberg2021patient}
Veronica Rotemberg, Nicholas Kurtansky, Brigid Betz-Stablein, Liam Caffery,
  Emmanouil Chousakos, Noel Codella, Marc Combalia, Stephen Dusza, Pascale
  Guitera, David Gutman, et~al.
\newblock A patient-centric dataset of images and metadata for identifying
  melanomas using clinical context.
\newblock {\em Scientific data}, 8(1):1--8, 2021.

\bibitem{streaks}
M Sadeghi, Tim~K. Lee, D McLean, H Lui, and M.~Stella Atkins.
\newblock Detection and analysis of irregular streaks in dermoscopic images of
  skin lesions.
\newblock {\em IEEE Transactions on medical imaging}, 32 (5):849--861, 2013.

\bibitem{pn}
M. Sadeghi, M. Razmara, Tim~K. Lee, and M.~Stella Atkins.
\newblock A novel method for detection of pigment network in dermoscopic images
  using graphs.
\newblock {\em Computerized Medical Imaging and Graphics}, 35 (2):137--143,
  2011.

\bibitem{Shafaei-OD-Test}
Alireza Shafaei, Mark Schmidt, and James Little.
\newblock {A Less Biased Evaluation of Out-of-distribution Sample Detectors}.
\newblock In {\em BMVC}, 2019.

\bibitem{SoenksenUD2021}
Luis~R. Soenksen, Timothy Kassis, Susan~T. Conover, Berta Marti-Fuster,
  Judith~S. Birkenfeld, Jason Tucker-Schwartz, Asif Naseem, Robert~R. Stavert,
  Caroline~C. Kim, Maryanne~M. Senna, José Avilés-Izquierdo, James~J.
  Collins, Regina Barzilay, and Martha~L. Gray.
\newblock Using deep learning for dermatologist-level detection of suspicious
  pigmented skin lesions from wide-field images.
\newblock {\em Science Translational Medicine}, 13(581):eabb3652, Feb. 2021.

\bibitem{lancet}
Philipp Tschandl, Noel Codella, Bengü~Nisa Akay, Giuseppe Argenziano, Ralph~P
  Braun, Horacio Cabo, David Gutman, Allan Halpern, Brian Helba, Rainer
  Hofmann-Wellenhof, Aimilios Lallas, Jan Lapins, Caterina Longo, Josep
  Malvehy, Michael~A Marchetti, Ashfaq Marghoob, Scott Menzies, Amanda Oakley,
  John Paoli, Susana Puig, Christoph Rinner, Cliff Rosendahl, Alon Scope,
  Christoph Sinz, H~Peter Soyer, Luc Thomas, Iris Zalaudek, and Harald Kittler.
\newblock Comparison of the accuracy of human readers versus machine-learning
  algorithms for pigmented skin lesion classification: an open, web-based,
  international, diagnostic study.
\newblock {\em The Lancet Oncology}, 20 (7):938--947, 2019.

\bibitem{ham10000}
P. Tschandl, C. Rosendahl, and H. Kittler.
\newblock The {HAM}10000 dataset, a large collection of multi-source
  dermatoscopic images of common pigmented skin lesions.
\newblock {\em Sci. Data}, 5, 2018.

\bibitem{metaEnsemble}
Apoorv Vyas, Nataraj Jammalamadaka, Xia Zhu, Dipankar Das, Bharat Kaul, and
  Theodore~L Willke.
\newblock Out-of-distribution detection using an ensemble of self supervised
  leave-out classifiers.
\newblock In {\em Proceedings of the European Conference on Computer Vision
  (ECCV)}, pages 550--564, 2018.

\bibitem{welch2021}
H~Gilbert Welch, Benjamin~L Mazer, and Adewole~S Adamson.
\newblock The rapid rise in cutaneous melanoma diagnoses.
\newblock {\em The New England journal of medicine}, 384(1):72--79, 2021.

\bibitem{paul2011}
P Wighton, Tim~K. Lee, H Lui, D.I. McLean, and M.S. Atkins.
\newblock Generalizing common tasks in automated skin lesion diagnosis.
\newblock {\em IEEE Transactions on Information Technology in Biomedicine}, 15
  (4):622--629, 2011.

\bibitem{msk}
Zachary~J. Wolner, Oriol Yelamos, Konstantinos Liopyris, Tova Rogers,
  Michael~A. Marchetti, and Ashfaq~A. Marghoob.
\newblock Enhancing skin cancer diagnosis with dermoscopy.
\newblock {\em Dermatol Clin.}, 35(4):417–437, 2017.

\bibitem{yap}
J Yap, W Yolland, and P Tschandl.
\newblock Multimodal skin lesion classification using deep learning.
\newblock {\em Exp Dermatol.}, 27:1261–1267, 2018.

\end{thebibliography}

\end{document}